%% file: main.tex
\pdfoutput=1

\documentclass[11pt]{article}

\usepackage[final]{acl}

\usepackage{times}
\usepackage{latexsym}

\usepackage[T1]{fontenc}

\usepackage[utf8]{inputenc}
\usepackage{microtype}
\usepackage{inconsolata}
\usepackage{graphicx}

\usepackage{lipsum}
\usepackage{threeparttable}
\usepackage{arydshln}
\usepackage{booktabs}
\usepackage{multirow}
\usepackage{algorithm}       
\usepackage{algpseudocode}     
\usepackage{amsmath}        
\usepackage{amssymb}        
\usepackage{enumerate}
\usepackage[shortlabels]{enumitem}

\input{macros}

\title{KG-CRAFT: Knowledge Graph-based Contrastive Reasoning with LLMs for Enhancing Automated Fact-checking}

\author{
 \textbf{V\'{i}tor N. Louren\c{c}o\textsuperscript{1,2}},
 \textbf{Aline Paes\textsuperscript{1}},
 \textbf{Tillman Weyde\textsuperscript{3}},
 \textbf{Audrey Depeige\textsuperscript{2}},
 \textbf{Mohnish Dubey\textsuperscript{2}}
\\
\\
 \textsuperscript{1}Universidade Federal Fluminense,
 \textsuperscript{2}Amazon,
 \textsuperscript{3}City St George's, University of London
\\
 \small{
   \textbf{Correspondence:} \href{mailto:vitorlourenco@id.uff.br}{vitorlourenco@id.uff.br}, \href{mailto:vitornl@amazon.co.uk}{vitornl@amazon.co.uk}
 }
}

\begin{document}
\maketitle
\begin{abstract}
Claim verification is a core component of automated fact-checking systems, aimed at determining the truthfulness of a statement by assessing it against reliable evidence sources such as documents or knowledge bases.
This work presents \name{}, a method that improves automatic claim verification by leveraging large language models (LLMs) augmented with contrastive questions grounded in a knowledge graph. 
\name{} first constructs a knowledge graph from claims and associated reports, then formulates contextually relevant contrastive questions based on the knowledge graph structure. 
These questions guide the distillation of evidence-based reports, which are synthesised into a concise summary that is used for veracity assessment by LLMs. 
Extensive evaluations on two real-world 
datasets (LIAR-RAW and RAWFC) demonstrate that our method achieves a new state-of-the-art in predictive performance. 
Comprehensive analyses validate in detail the effectiveness of our knowledge graph-based contrastive reasoning approach in improving LLMs' fact-checking capabilities.
\end{abstract}

\section{Introduction}
\input{secs/1-intro}


\section{Related Work}
\input{secs/2-background}

\section{Knowledge Graph-based Contrastive Reasoning}\label{sec:3kgcraft}

\begin{figure*}[t]
    \centering
    \includegraphics[trim={0 4.3cm 0 0.7cm},clip, width=\textwidth]{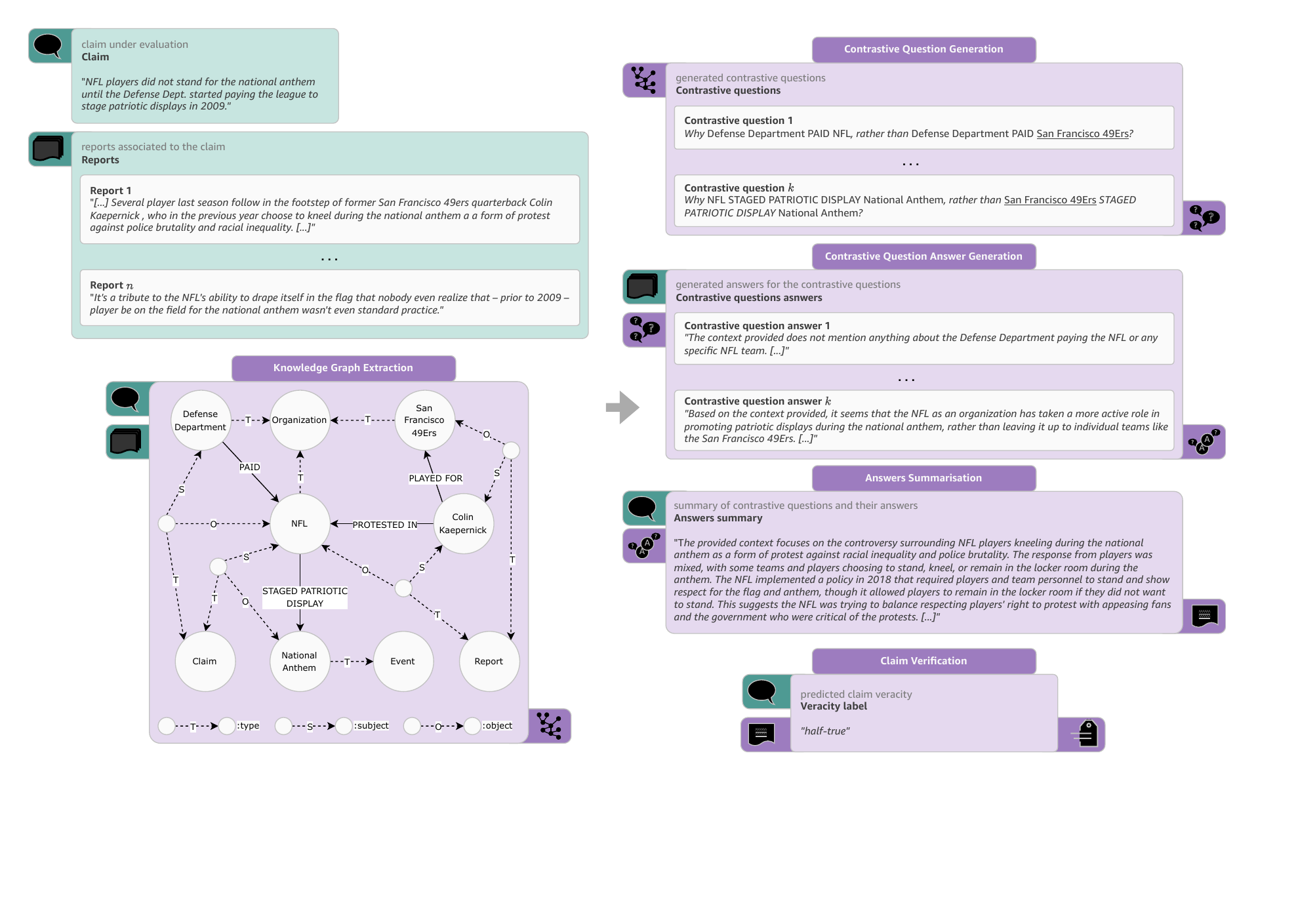}
    \caption{Overview of the \name{} framework for automated fact-checking. The method comprises three main phases: (1) knowledge graph extraction from the claim and associated reports, (2) contrastive reasoning, including contrastive question formulation, answer generation, and answer summarisation, and (3) claim veracity prediction.}
    \label{fig:framework}
\end{figure*}

\input{secs/3-methodology}

\section{Experiments and Results}
\input{secs/4-results}

\section{Conclusion}
\input{secs/5-conclusion}


\section*{Limitations}
\input{secs/6-limitations}

\section*{Ethical Considerations}
Automated fact-checking with LLMs raises important ethical concerns, particularly due to their ability to produce fluent yet incorrect or fabricated outputs. When used to synthesise evidence or generate contrastive questions, such errors can mislead users or amplify misinformation. Condensing large evidence sets also involves decisions about what to include or omit; if not carefully managed, this can result in oversimplification or the exclusion of critical context, compromising factual accuracy. Moreover, whilst contrastive reasoning may improve interpretability, the overall decision-making process remains difficult to audit, with limited traceability from evidence to claim and limited justification for model outputs. Automated fact-checking systems may also be misinterpreted as authoritative, reinforcing confirmation bias. It is therefore essential to clearly communicate model limitations and maintain human oversight. We emphasise the importance of fairness-aware design, transparent evidence attribution, and rigorous human evaluation to mitigate these risks.

\subsection*{Use of Generative AI}
During the preparation of this work, the authors used the Claude family models and the Amazon Nova family models to grammar and spelling check. After using these tools/services, the authors reviewed and edited the content as needed and took full responsibility for the publication's content. 

\section*{Acknowledgments}
The second author thanks the support of FAPERJ - \textit{Fundação Carlos Chagas Filho de Amparo à Pesquisa do Estado do Rio de Janeiro}, processes SEI-260003/002930/2024, SEI-260003/000614/2023, CNPq (National Council for Scientific and Technological Development), (grant \#307088/2023-5) and the National Institutes of Science and Technology (INCT): IAIA (grant \#406417/2022-9), TILD-IAR (grant \#408490/2024-1), and IAPROBEM (grant \#408589/2024-8).

\bibliography{main}

\appendix

\include{secs/a5-background}

\include{secs/a1-implementation}

\include{secs/a3-exp}

\include{secs/a2-prompts}

\end{document}

%% file: macros.tex
\usepackage{soul}
\usepackage{xcolor}
\usepackage{hyperref}
\usepackage{cleveref}
\usepackage[most]{tcolorbox}
\usepackage{lmodern}
\usepackage{parskip}
\usepackage{geometry}
\usepackage{xparse}

\sethlcolor{yellow}

\NewDocumentCommand{\promptbox}{O{} m m}{%
  \begin{tcolorbox}[
    enhanced jigsaw, breakable,
    colback=white,
    colframe=gray!60,
    coltitle=white,
    colbacktitle=black,
    fonttitle=\sffamily\bfseries,
    title=#2,
    boxrule=0.6pt,
    arc=2mm,
    left=6mm, right=6mm,
    top=3mm, bottom=3mm,
    boxed title style={left=6mm, right=6mm, top=2mm, bottom=2mm},
    before skip=8pt plus 2pt minus 1pt,
    after  skip=10pt plus 2pt minus 1pt,
    shadow={0pt}{0pt}{6pt}{black!10},
  ]
  {\ttfamily\csname #1\endcsname \raggedright #3}
  \end{tcolorbox}
}

\algnewcommand\algorithmicforeach{\textbf{for each}}
\algdef{S}[FOR]{ForEach}[1]{\algorithmicforeach\ #1\ \algorithmicdo}

\crefname{chapter}{Chapter}{Chapters}
\crefname{section}{Section}{Sections}
\crefname{subsection}{Section}{Sections}
\crefname{subsubsection}{Section}{Sections}
\crefname{figure}{Figure}{Figures}
\crefname{table}{Table}{Tables}
\crefname{equation}{Equation}{Equations}
\crefname{subfigure}{Figure}{Figures}
\crefname{algorithm}{Algorithm}{Algorithms}
\crefname{subsubfigure}{Figure}{Figures}
\crefname{appendix}{Appendix}{Appendices}

\definecolor{cb-black}{RGB}{0,0,0}
\definecolor{cb-orange}{RGB}{230,159,0}
\definecolor{cb-skyblue}{RGB}{86,180,233}
\definecolor{cb-bluishgreen}{RGB}{0,158,155}
\definecolor{cb-yellow}{RGB}{240,228,66}
\definecolor{cb-blue}{RGB}{0,114,178}
\definecolor{cb-vermillion}{RGB}{213,94,0}
\definecolor{cb-reddishpurple}{RGB}{204,121,167}
\definecolor{darkgreen}{rgb}{0.0, 0.2, 0.13}

\newcommand{\eg}{\emph{e.g.}}
\newcommand{\ie}{\emph{i.e.}}

\newcommand{\name}{KG-CRAFT}
\newcommand{\namemodel}[1]{KG-CRAFT$_{#1}$}

%% file: secs/1-intro.tex


The digital transformation has reshaped how society consumes and shares information, posing new challenges to information integrity~\cite{haider2022paradoxes}. \textit{The Reuters Institute Digital News Report 2024} highlights how social media has fragmented the news ecosystem~\cite{newman2024reuters}. 
Despite expanding access and engagement, this shift has also fuelled the spread of misinformation~\cite{valenzuela2019paradox}.

Misinformation is particularly concerning in high-stakes contexts such as elections and public health crises, where it can cause serious societal harm.
Consequently, the demand for more effective and scalable fact-checking methods has driven the emergence of Automated Fact-Checking~(AFC) systems~\cite{Zhou2020,alam2022,Guo2022,eldifrawi2024automated}.


AFC is designed to assess the veracity of claims by retrieving, analysing, and reasoning over relevant evidence from reliable sources~\cite{wu2025systematic}. 
Early approaches relied on classification and evidence retrieval pipelines~\cite{shu2019defend,kotonya2020explainable,atanasova2020generating}. 
Whilst the integration of relational structures and knowledge bases~\cite{lourenco2022modality,tillman2022icwsm,lourenco2025bracis,huang2025graphfc,chen2025graphcheck}, and hierarchical architectures~\cite{yang2022coarse} improved performance, these methods lacked the scalability and adaptability that LLMs later demonstrated. 
Recent LLM-based approaches have achieved significant advances in AFC through the integration of external knowledge~\cite{cheung2023factllama,guo2023interpretable} and the introduction of retrieval mechanisms~\cite{singal2024evidence,zhang2024reinforcement,yue2024retrieval}. However, these solutions often lack structured reasoning mechanisms~\cite{liu2024teller}, which can lead to unreliable verification processes. 
Contrastive reasoning~\cite{jacovi2021contrastive,paranjape2021contrastive} has demonstrated effectiveness in enhancing model interpretability and decision-making, yet its application to fact-checking remains underexplored.

To address the aforementioned open challenges and ultimately enhance AFC capabilities, we propose a method for improving \emph{claim verification} within the AFC pipeline. 
We focus on claim verification in a bounded context, where each claim is accompanied by a predefined set of associated reports. This setting reflects numerous domains, \eg{}, legal document review~\cite{Zheng2021casehold}, financial auditing~\cite{zhu2021tatqa}, and scientific peer review~\cite{wadden2020scifact}, where analysis is restricted to a specific corpus. 
Consequently, the core challenge shifts from open-domain evidence retrieval to reasoning effectively over the available information to determine a claim’s veracity. 
Specifically, we propose \emph{Knowledge Graph-based Contrastive Reasoning for Automated FacT verification}~(\name{}).

Motivated by findings in both cognitive science and natural language processing~\cite{DBLP:conf/naacl/SchusterFB21,DBLP:conf/chi/BucincaSPDG25}, we leverage contrastive reasoning for our automated fact-checking task. 
Prior work highlights that verification requires distinguishing whether a claim is supported or contradicted~\cite{thorne2018fever,aly2021feverous}, and that contrastive explanations align more closely with human reasoning~\cite{miller2019}. Moreover, contrastive learning methods have proven effective in enforcing meaningful semantic distinctions~\cite{chen2020simclr,gao2021simcse}. 
We therefore introduce contrastive questions into the fact-checking pipeline, encouraging models not only to assess whether evidence supports a claim but also to explicitly consider alternatives, thereby promoting more robust verification.

However, generating meaningful contrasts from unstructured text alone is a non-trivial challenge. 
Without a structured representation of the underlying facts and their relationships, contrastive questions may focus on superficial linguistic differences rather than semantically significant distinctions, leading to less informative or even arbitrary contrasts~\cite{DBLP:conf/ijcai/BhattacharjeeK022}. 
By explicitly encoding entities and their semantic relations, a knowledge graph~(KG) provides a structured means of identifying candidates for contrast~\cite{liu2019kg}. This structure guides the model to formulate questions that explore meaningful conceptual distinctions (\eg{}, contrasting one entity with another of the same type), ensuring that the reasoning process is both more robust and more aligned with human-like. 
This approach helps to ensure that contrasts are grounded in verification-relevant relationships rather than being purely text-driven.

To implement this structured contrastive approach, \name{} first decomposes each claim and its associated reports into entities and their relationships to construct the KG.
After, it formulates and selects contrastive questions based on the KG structure, aiming to maximise both diversity and contextual relevance to the claim. 
The set of questions is then answered using the input reports, generating a new, contextually relevant, evidence-based information set. Inspired by the scalability and demonstrated performance of LLMs in automated fact-checking~\cite{cheung2023factllama,wang2024explainable,xiong2025delphi}, this new set is consolidated into a concise summary representing a distilled version of the input reports, which is then used to assess the veracity of the claim.


The main contributions of this work are: \textit{(i)} a novel Knowledge Graph-based Contrastive Reasoning method that enhances LLM capabilities in AFC; \textit{(ii)} state-of-the-art performance on two real-world fact-checking datasets; and \textit{(iii)} a comprehensive ablation study analysing the proposed components for AFC.

%% file: secs/2-background.tex

This section situates our contribution within prior work on AFC, and LLM-based claim verification. 
Foundational concepts are deferred to \cref{sec:abg}, which provides formal definitions and notation for (i) \emph{contrastive explanations and reasoning}, and (ii) \emph{knowledge graph construction with LLMs}; these concepts are developed in greater detail in the appendix. 
\cref{sec:abg} also presents an extended review of related work, including further discussion of KG-based approaches to fact-checking.

\textbf{Non-generative Automated Fact-checking\label{sec:2ml}}
Classical AFC models encode claims and documents using text embeddings, and verify them via supervised classifiers.
Notable systems include dEFEND, which employs sentence–comment co-attention for news and user comments~\cite{shu2019defend}; SBERT-FC, which introduced the PubHealth dataset, and an explainability analysis~\cite{kotonya2020explainable}; and GenFE/GenFE-MT, which jointly optimise veracity prediction and explanation generation~\cite{atanasova2020generating}. 
CofCED proposes a hierarchical encoder with cascaded evidence selectors for multi-source reports~\cite{yang2022coarse}. 
Incorporating KGs into pretrained models (\eg{}, via Wikidata) improves accuracy, especially for political claims~\cite{tillman2022icwsm}; a recent survey comprehensively reviews KG-based AFC~\cite{qudus2025kgfc}.


\textbf{Fact-checking Using LLMs\label{sec:2llm}}
LLMs have become central to AFC, yet their reliability remains constrained by training coverage and hallucinations~\cite{wang2024factuality,augenstein2024factuality}.
Early systems augment LLMs with structure or external evidence: FactLLaMA couples instruction-following with retrieval~\cite{cheung2023factllama}; IKA builds example graphs for verification and explanation~\cite{guo2023interpretable}; TELLER integrates human expertise with LLM reasoning~\cite{liu2024teller}; defence-style frameworks partition evidence into competing narratives for robust verification~\cite{wang2024explainable}; and CorXFact models claim–evidence correlations~\cite{tan2025improving}. 
Retrieval-augmented generation~(RAG), ranging from basic RAG pipelines~\cite{singal2024evidence}, to retrieval optimised with fine-grained feedback~\cite{zhang2024reinforcement}, and architectures targeting evidence retrieval plus contrastive argument synthesis~\cite{yue2024retrieval}, has become an increasingly prominent approach. 
Other recent directions include iterative verification for scalability (FIRE)~\cite{xie2025fire}, and the handling of zero-day manipulations via real-time context retrieval~\cite{meng2025detecting}.

%% file: secs/3-methodology.tex


We introduce a novel approach that leverages knowledge graphs to fuel contrastive reasoning and enhance LLM fact-checking capabilities: \emph{Knowledge Graph-based Contrastive Reasoning for Automatic FacT Verification}~(\name{}).
Our work focuses on the claim verification component of automated fact-checking, integrating contrastive reasoning into the verification process through the use of structured evidence to generate contextually relevant contrastive queries, thereby guiding more accurate claim classification.
We begin with the claim verification task formulation.


\textbf{Problem Statement}
Let $\mathcal{C}$ be a claim with a set of associated \emph{reports}
$\mathcal{R}_{\mathcal{C}}=\{r_i\}_{i=1}^{|\mathcal{R}_{\mathcal{C}}|}$, where each report
$r_i=(s_{i,1},\ldots,s_{i,|r_i|})$ is a sequence of sentences (\ie{}, a document).
Optionally, sentence-level \emph{evidence} annotations are given as a set of
indices $\varepsilon_{\mathcal{C}}\subseteq\{(i,j)\}$; if unavailable, set
$\varepsilon_{\mathcal{C}}=\varnothing$. The objective is to predict a veracity label
$\mathcal{V}_\mathcal{C}\in\mathcal{Y}$, $|\mathcal{Y}| \ge 2$ via a verifier
$f_\theta:\ (\mathcal{C},\mathcal{R}_{\mathcal{C}})\mapsto \mathcal{V}_\mathcal{C}$.
Evidence $\varepsilon_{\mathcal{C}}$ (when present) is used for analysis but is not required by the formulation.

Next, we present the components of \name{}, depicted in \cref{fig:framework}: knowledge graph construction from the textual input (the claim and its associated reports) (\cref{sec:3kge}); contrastive reasoning, comprising contrastive question generation, answer generation, and prompt-based answer summarisation (\cref{sec:3cr}); and claim veracity verification (\cref{sec:3cv}).


\subsection{Knowledge Graph Extraction}\label{sec:3kge}
The first phase of \name{} extracts entities and relationships from $\mathcal{C}$ and
$\mathcal{R}_{\mathcal{C}}$ and uses them to construct a knowledge representation of the input. We draw inspiration from prior work~\cite{zhu2024llms,zhang2024extract} and leverage LLMs for knowledge graph construction. Through phased few-shot prompting, we instrument the LLM to first identify entities $\mathcal{E}$, then label them according to their conceptual categories $\mathbb{C}$, to give them semantic meaning and enable disambiguation. Next, we identify the relationships $\mathcal{R}$ that relate the identified entities (prompt details in \cref{sec:akgep}), forming a set of triples \(\mathcal{T} \subseteq \mathcal{E} \times \mathcal{R} \times \mathcal{E}\). 
The unified resulting sets constitute the input knowledge graph $\mathcal{G}_{{\mathcal{C}},\mathcal{R}_{\mathcal{C}}} = (\mathcal{E}, \mathcal{R}, \mathcal{T}, \mathbb{C})$.

\subsection{Contrastive Reasoning}\label{sec:3cr}
The second phase of \name{} uses \(\mathcal{G}_{{\mathcal{C}},\mathcal{R}_{\mathcal{C}}}\) to formulate and answer contrastive questions. 
It consists of: 
\textit{(i)} formulating questions that contrast the claim’s facts ($\mathcal{T}_{\text{claim}} \subseteq \mathcal{T}$) with the reports’ facts ($\mathcal{T} - \mathcal{T}_{\text{claim}}$); 
\textit{(ii)} answering the formulated questions using the reports ($\mathcal{R}_\mathcal{C}$); and 
\textit{(iii)} summarising the question-answer pairs into a single self-contained paragraph.

\textbf{Contrastive Question Formulation\label{sec:3cqf}}
\cref{alg:fcq} describes the process to formulate contrastive questions.
Given $\mathcal{G}_{{\mathcal{C}},\mathcal{R}_{\mathcal{C}}}$, the claim-specific triples $\mathcal{T}_{\text{claim}}$, and a maximum number $\texttt{K}$ of desired questions, the algorithm creates the $\texttt{K}$ most relevant and diverse contrastive questions. 
For each triple in the claim (consisting of $head$ entity, relation, and $tail$ entity), it first identifies the entity categories $h_c$ and $h_t$ $\in \mathbb{C}$ of the head and tail entities, respectively(\cref{alg:fcq},~ll.6-8). 
Then, it creates two sets of contrastive entities: $H_{contr}$, containing alternative head entities of category $h_c$ and $T_{contr}$, with alternative tail entities of category $h_t$ (\cref{alg:fcq},~ll.9-10). Using these sets, the algorithm generates questions by replacing either the original head or tail entity whilst maintaining the relation, following the pattern ``Why [\textit{original head}] rather than [alternative]?'' or ``Why [alternative] rather than [\textit{original tail}]?'' (\cref{alg:fcq},~ll.11-14).

\input{resources/algo-cqgen}

To ensure that the formulated questions are individually relevant and collectively diverse, we adopt a ranking strategy based on Maximal Marginal Relevance~(MMR) (\cref{alg:fcq},~l.20). 
For that, we first compute embeddings of the questions $\mathcal{Q}_{\text{Em}} = \{{\textsc{Embedding}(q_i)} \mid q_i \in \mathcal{Q}\}_{i=1}^{\mid\mathcal{Q}\mid}$, followed by the pairwise similarity matrix across all embeddings. 
The initial query embedding $\texttt{q}_\theta$ is selected as the embedding with the highest average similarity to all others, thereby ensuring a representative starting point.

We then iteratively construct the ranked set $\mathcal{Q}_{ranked}$ using the MMR procedure. At each iteration, the next embedding $q_i$ is selected to maximise
\begin{equation}\label{eq:mmr}
\small
\texttt{q}_i = \mathop{\arg\max}_{\texttt{q} \in \mathcal{Q}_{\text{Em}} \setminus \mathcal{Q}_{ranked}} \left[ \text{Sim}(\texttt{q}, \texttt{q}_\theta) - \max_{\texttt{q}' \in \mathcal{Q}_{ranked}} \text{Sim}(\texttt{q}, \texttt{q}') \right],
\end{equation}
where $\text{Sim}(\cdot,\cdot)$ is the cosine distance, and $\mathcal{Q}_{ranked}$ is the set of already selected embeddings. The first term promotes relevance to the initial query $\texttt{q}_\theta$, whilst the second penalises redundancy with respect to $\mathcal{Q}_{ranked}$. The process is repeated until all candidate questions are ranked into $\mathcal{Q}_{ranked}$.
Finally, the algorithm returns the top~\texttt{K} questions $\mathcal{Q}_{ranked}^\texttt{K}$ (\cref{alg:fcq},~ll.17-18). 

\textbf{Contrastive Question Answer Generation\label{sec:3cqag}}
The second step of the process answers the previously formulated contrastive questions $\mathcal{Q}_{ranked}^\texttt{K}$ through a structured query-response prompt mechanism $p_{\text{ag}}$ (available in \cref{sec:aagp}). 
An LLM analyses and generates information from the claim-associated reports -- prompting it to answer independently the contrastive questions based on the set of reports $\mathcal{R}_\mathcal{C}$ associated with the claim. 
This process generates a set of answers $\tilde{\mathcal{A}} = \{LLM_{p_{\text{ag}}}(q, \mathcal{R}_\mathcal{C}) \mid q \in \mathcal{Q}_{ranked}^\texttt{K}\}$, where each answer is directly derived from the reports, maintaining traceability between claim, reports, contrastive questions, and answers. Our goal by highlighting the contrastive elements in the generated answers is to highlight the key evidence supporting the claim's veracity during the reasoning process.

\textbf{Answers Summarisation\label{sec:3as}}
The final step of the Contrastive Reasoning process involves aggregating the question-answer pairs into a concise, evidence-based summary. From the claim $\mathcal{C}$ and the paired contrastive questions and answers from $\mathcal{Q}_{ranked}^\texttt{K}$ and $\tilde{\mathcal{A}}$, respectively, we prompt $p_{as}$ (\cref{sec:aasp}) to an LLM to generate a concise paragraph that relates all contrastive question-answer pairs. This summarisation step, represented as $A_{\mathcal{C}} = LLM_{p_{as}}(\mathcal{C}, \{(q_i, a_i) \mid q_i \in \mathcal{Q}_{ranked}^\texttt{K},\, a_i \in \tilde{\mathcal{A}}\})$, ensures that key contrasting elements and supporting evidence are presented in a structured summary, producing a distilled source of information. The resulting summary $A_{\mathcal{C}}$ emphasises key contrasting facts whilst abstracting non-essential information, preserving semantic links between critical evidence and the claim to create a focused source of verification.

\subsection{Verification of Claim Veracity\label{sec:3cv}}
The last phase of \name{} assesses the claim's veracity $\mathcal{V}_\mathcal{C}$. 
For that, we prompt $p_\text{cv}$ (see \cref{sec:acvvp}) an LLM, containing the original claim $\mathcal{C}$ and the produced summary $A_{\mathcal{C}}$ as the only source of evidence, mitigating potential noise in the original reports $\mathcal{R}_\mathcal{C}$. 
The prompt $p_\text{cv}$ also includes the possible labels and their descriptions. 
In this way, the LLM acts as a classifier to infer the claim's veracity, represented as $\mathcal{V}_\mathcal{C} = LLM_{p_\text{cv}}(\mathcal{C}, A_{\mathcal{C}})$, ensuring that the veracity assessment is based on the distilled evidence produced through our Contrastive Reasoning method.

%% file: resources/algo-cqgen.tex
\begin{algorithm}[t]
\small
\caption{Define Contrastive Questions}
\label{alg:fcq}
\begin{algorithmic}[1]

\algblock{Function}{EndFunction}
\algblock{Procedure}{EndProcedure}

\Require
    \State $\mathcal{G} = (\mathcal{E}, \mathcal{R}, \mathcal{T}, \mathbb{C})$\!\hspace{0.5em}{$\triangleright$ Claim and reports KG; entities $\mathcal{E}$, relations $\mathcal{R}$, triples $\mathcal{T}$, and entities' classes $\mathbb{C}$}

    \State $\mathcal{T}_{\text{claim}}$\!\hspace{0.5em}{$\triangleright$ Claim's extracted triples; $\mathcal{T}_{\text{claim}} \subseteq \mathcal{T}$}

    \State $\texttt{K}$ \Comment{Maximum \# of Contrastive Questions}
\Ensure
    \State $\mathcal{Q}_{ranked}^\texttt{K}$\!\hspace{0.5em}{$\triangleright$ Set of top $\texttt{K}$ most relevant and diverse contrastive questions for a claim and its reports}
    
\Statex

\State $\mathcal{Q} \gets \emptyset$

\ForEach{$\texttt{t} = \{h,r,t\} \in \mathcal{T}_{\text{claim}}$}
    \State $h_{c} \gets \tau(h)$ \Comment{where $\tau : \mathcal{E} \rightarrow \mathbb{C}$}
    \State $t_{c} \gets \tau(t)$
    
    \State $H_{contr} \gets \{h \mid h \in \mathcal{E}, \tau(h) = h_{c} \} \setminus \{h\}$\\\Comment{Set of entities of the same class as $head$}
    \State $T_{contr} \gets \{t \mid t \in \mathcal{E}, \tau(t) = t_{c} \} \setminus \{t\}$\\\Comment{Set of entities of the same class as $tail$}
    
    

        \ForEach{$h' \in H_{contr}\;, t' \in T_{contr}$}
        \State $q_h \gets$ $\textsc{\small{FormulateQuestion}}(\texttt{t}, h')$
        \State $q_t \gets$ $\textsc{\small{FormulateQuestion}}(\texttt{t}, t')$
        \State $\mathcal{Q} \gets \mathcal{Q} \cup \{q_h, q_t\}$
    \EndFor
\EndFor

\State $\mathcal{Q}_{ranked} \gets \textsc{ReRank} (\mathcal{Q})$ \Comment{Rank contrastive questions based on \cref{eq:mmr}}

\State $\mathcal{Q}_{ranked}^\texttt{K} \gets (\mathcal{Q}_{ranked})_{1:K}$ \Comment{Select first K elements from ranked set}

\Statex
\State \Return $\mathcal{Q}_{ranked}^\texttt{K}$
\end{algorithmic}
\end{algorithm}



%% file: secs/4-results.tex
This section is guided by the following research questions: 
\textbf{RQ1} How effective is \name{} in claim verification compared to state-of-the-art methods? 
\textbf{RQ2} How beneficial are KG-based contrastive questions compared to purely LLM-generated contrastive questions? 
\textbf{RQ3} What is the effect of the number of contrastive questions $\texttt{K}$ on claim verification with \name{}? 
\textbf{RQ4} How effective is \name{} with Small Language Models~(SLMs) compared to LLMs? 
To answer these questions, we evaluate \name{} on two real-world fact-checking benchmarks and compare the results with baseline methods, and perform several ablation studies. Additional experiments and ablation studies can be found in~\cref{sec:aexp}.

\subsection{Experimental Settings}\label{sec:4es}

\textbf{Datasets} We evaluated the proposed approach on two publicly available fact-checking datasets: LIAR-RAW and RAWFC, (refer \cref{sec:ads} for datasets statistics). 
LIAR-RAW~\cite{yang2022coarse} extends LIAR-PLUS~\cite{alhindi2018evidence} and contains fine-grained claims from Politifact\footnote{\url{www.politifact.com}, \footnotemark\url{www.snopes.com}} with six veracity classes (\texttt{pants-fire}, \texttt{false}, \texttt{barely-true},
\texttt{half-true}, \texttt{mostly-true}, \texttt{true}) along with their relevant reports. RAWFC~\cite{yang2022coarse} contains claims collected from Snopes\footnotemark[2]~on various topics with three veracity classes (\texttt{false}, \texttt{half}, \texttt{true}) and their associated reports retrieved using claim keywords.

\textbf{Comparisons} We compare \name{} against other methods in three categories: \textit{Traditional} -- methods that do not use LLMs --, \textit{Na\"{i}ve LLM} -- direct application of LLMs without specialised prompts or reasoning strategies --, and \textit{Specialised LLM} -- methods that use LLMs with significant adaptations, such as tuning or prompt engineering, or as part of complex architectures. 
Traditional approaches encompass dEFEND~\cite{shu2019defend},
SBERT-FC~\cite{kotonya2020explainable}, GenFE and GenFE-MT~\cite{atanasova2020generating}, and CofCED~\cite{yang2022coarse}. 
Na\"{i}ve LLM approaches encompass Llama~2~7B and ChatGPT~3.5~Turbo~\cite{wang2024explainable}, Claude~3.5~Sonnet, Claude~3.7~Sonnet, and Llama~3.3~70B prompted with the claim and related reports to verify the veracity of the claim. 
The Specialised LLM approaches encompasses FactLLAMA and FactLLAMA\textsubscript{know}~\cite{cheung2023factllama}, and L-Defense\textsubscript{LLAMA2} and L-Defense\textsubscript{ChatGPT}~\cite{wang2024explainable}. FactLLAMA is a Low-Rank Adaptation~(LoRA)~\cite{hu2022lora} fine-tuned Llama~2~7B, and FactLLAMA\textsubscript{know} is the model augmented with external knowledge. L-Defense is a defence-based framework that leverages the wisdom of crowds to verify claim veracity using Llama~2~7B and ChatGPT~3.5~Turbo.

\textbf{Implementation Details} We extracted the KGs (\cref{sec:3kge}) of both datasets utilising Claude~3~Haiku. Further, \name{} is instantiated and evaluated (\cref{sec:4cve}) using Claude~3.5~Sonnet (\namemodel{C3.5}), Claude~3.7~Sonnet (\namemodel{C3.7}), and Llama~3.3~70B (\namemodel{L3.3}) (refer to Appendix~\ref{sec:aid} for more implementation details).

\textbf{Evaluation Metrics} We adopted standard classification metrics: precision~(Pr), recall~(Re), and F1-score~(F1).
For all metrics, higher values indicate better performance.

\subsection{Claim Verification Evaluation}\label{sec:4cve}

\input{resources/tab-verification}

We first address \textbf{RQ1} by evaluating the performance of \name{} in verifying claim veracity. 
As shown in Table~\ref{tab:verification}, our method -- in the instances \namemodel{C3.7} and \namemodel{L3.3} (all using five contrastive questions $\texttt{K}=5$) -- consistently outperforms all other methods on both datasets. 
\namemodel{C3.5} outperforms all comparator methods except for DelphiAgent\textsubscript{gpt-4o}) on the RAWFC dataset. 
\namemodel{L3.3} improves the F1-score by 44 percentage points~(pp) on the LIAR-RAW dataset and 13 pp on the RAWFC dataset, compared to the second best performing methods, L-Defense~\cite{wang2024explainable} and (DelphiAgent~\cite{xiong2025delphi}), respectively.
Compared to their Na\"{i}ve LLM counterparts, \namemodel{C3.5}, \namemodel{C3.7}, and \namemodel{L3.3}, show F1 performance gains of 32, 44, and 42 pp on LIAR-RAW and 11, 12, and 27 pp on RAWFC.

\subsection{Ablation Studies\label{sec:4as}}
To evaluate \name{}'s components and reveal its strengths and limitations, we conduct three ablation studies. 
We first compare KG-based and LLM-generated contrastive questions to evaluate the benefits of structured question formulation (\textbf{RQ2}). Then, we analyse the effect of varying the number $\texttt{K}$ of contrastive questions (\textbf{RQ3}). Finally, we assess whether \name{} boosts the performance of Small Language Models~(SLMs) by comparing their results to Claude~3.7~Sonnet and \namemodel{C3.7} (\textbf{RQ4}).

\subsubsection{Impact of Using LLM-generated Contrastive Questions}\label{sec:4llmcq}
To answer \textbf{RQ2}, we replaced the Contrastive Question Formulation component of \name{} (\cref{sec:3cqf}) by a few-shot prompt that, given the claim, reports, and examples, requests the LLM to generate $k=5$ contrastive questions (prompt details in \cref{sec:cqfp}). Results depicted on \cref{tab:llmcontrastive} show fact-checking F1-score, and macro and weighted AlignScore and RQUGE for the LIAR-RAW dataset.

\emph{AlignScore} is a metric based on a general function of information alignment to perform automatic factual consistency evaluation of text pairs~\cite{zha2023alignscore}. In our evaluation, we use AlignScore (here called macro AlignScore) to measure the information alignment between the text piece generated at the answer summarisation step (\cref{sec:3as}) with the original claim.
\emph{RQUGE} is a reference-free evaluation metric for generated questions, based on the corresponding context and answer~\cite{mohammadshahi2023rquge}. In our evaluation, we use RQUGE (here called macro RQUGE) to measure the acceptability score of a formulated contrastive question (\cref{sec:3cqf}), given a corresponding context (the claim) and an answer (the answer generated from the contrastive question~\cref{sec:3cqag}).
We further extend AlignScore and RQUGE (here called weighted AlignScore and weighted RQUGE) to weight their results by the distance of the predicted claim veracity class from the ground truth claim class. For that, we assigned each class an integer numerical value ranging from 1 (\texttt{pants-fire}) to 6 (\texttt{true}) and measured the mean squared error~(MSE) between the predicted class. After, we weight AlignScore and RQUGE macro scores using the obtained MSE (details of the AlignScore and RQUGE weighted scores are in \cref{sec:awalignscore} and \cref{sec:awrquge}, respectively).


\begin{table}[ht]
\centering
\small
\caption{Ablation study on the LIAR-RAW dataset comparing the quality of KG-based in contrast to LLM-based formulated contrastive questions. Metrics include Fact-checking F1-score, and macro (-M) and weighted (-W) values for AlignScore (AS) and RQUGE (RQ).}
\begin{tabular}{lcccc}
\toprule
\textbf{Metric} & \multicolumn{2}{c}{\textbf{\small{LLM-based}}} & \multicolumn{2}{c}{\textbf{\small{KG-based}}} \\
\cmidrule(lr){2-3} \cmidrule(lr){4-5}
& \textbf{\small{C3.5}} & \textbf{\small{L3.3}} & \textbf{\small{C3.5}} & \textbf{\small{L3.3}} \\
\midrule
\textbf{FC F1} & 27.79 & 29.68 & 60.99 & 73.87 \\
\midrule
\textbf{AS-M} & 30.96 & 29.51 & 41.71 & 40.32 \\
\textbf{AS-W} & 29.15 & 24.36 & 39.17 & 35.84 \\
\midrule
\textbf{RQ-M} & 2.04 & 1.92 & 2.02 & 1.95 \\
\textbf{RQ-W} & 1.50 & 1.43 & 1.67 & 1.64 \\
\bottomrule
\end{tabular}
\label{tab:llmcontrastive}
\end{table}

As depicted in~\cref{tab:llmcontrastive}, the contrastive questions generated by the LLMs, results in lower information alignment (AlignScore) between the answer summary $A_\mathcal{C}$ and claim $\mathcal{C}$ and lower question acceptability score (RQUGE) between formulated contrastive questions $Q^\texttt{K}_{ranked}$, generated answer $\tilde{A}$, and claim $\mathcal{C}$. As a consequence, the lower information alignment and question acceptability score impact the fact-checking capability, resulting in a lower fact-checking F1-score.

\subsubsection{Impact of the Number of Contrastive Questions}\label{sec:4kcq}

To address \textbf{RQ3}, we investigate how varying the number of contrastive questions $k$ affects the performance of \name{}. We conduct this analysis on a binary classification variant of LIAR-RAW, where the original six labels are mapped into two classes: \{\texttt{pants-fire}, \texttt{false}, \texttt{barely-true}\} as \texttt{false}, and \{\texttt{half-true}, \texttt{mostly-true}, \texttt{true}\} as \texttt{true} (details in \cref{sec:ads}). We evaluate \namemodel{C3.5} using $k \in \{1,3,5,7,10\}$ contrastive questions, with $\texttt{K}=5$ serving as our baseline for relative performance comparison.

\begin{figure}[htbp]
    \centering
    \includegraphics[width=0.925\linewidth]{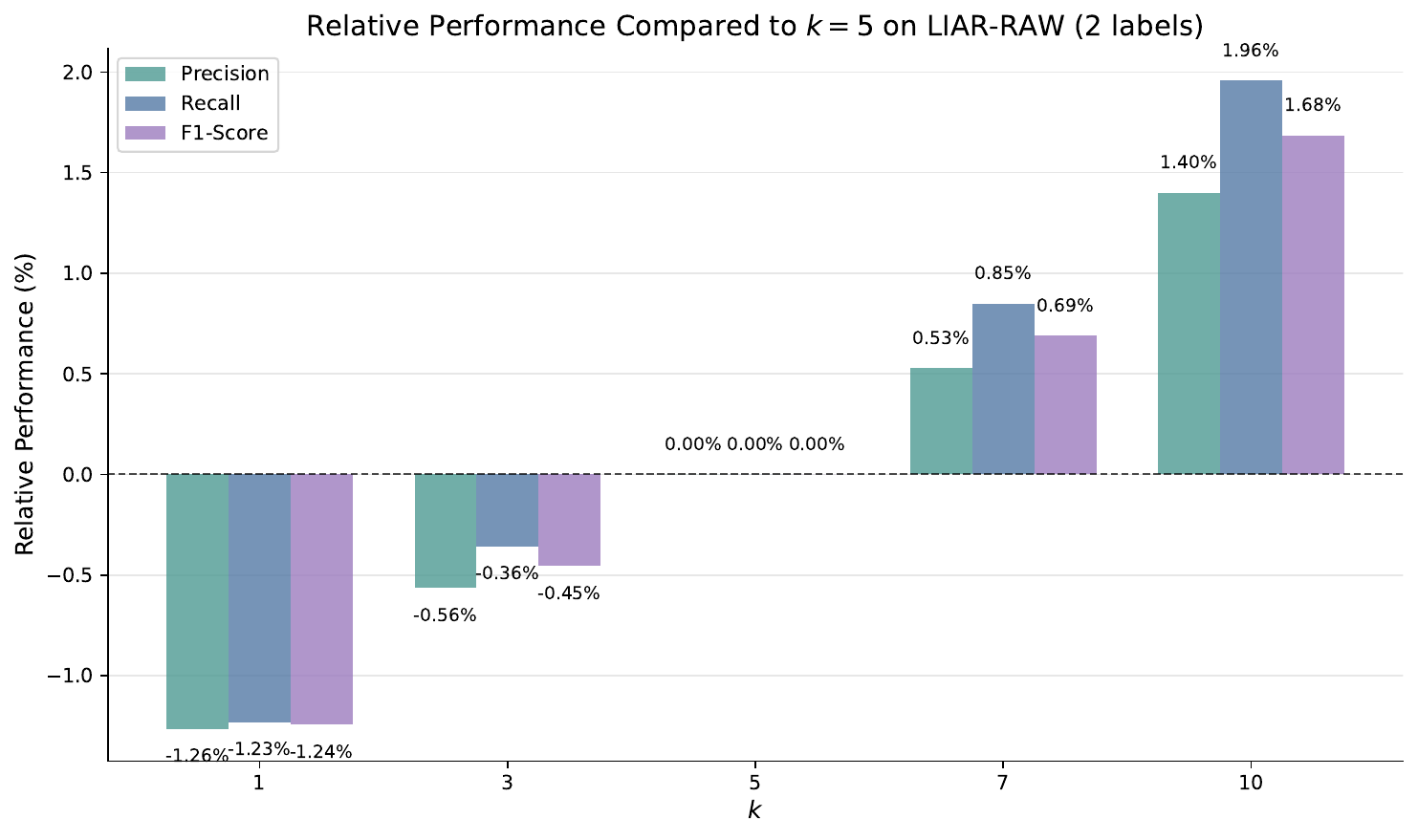}
    \caption{Impact of varying the number of contrastive questions $k$ on fact-checking performance.}
    \label{fig:topk}
\end{figure}

As depicted in Figure~\ref{fig:topk}, increasing $\texttt{K}$ generally improves performance metrics, but with progressively smaller gains.
Specifically, compared to $\texttt{K}=5$, using fewer questions ($\texttt{K}=1$ or $\texttt{K}=3$) leads to decreased performance, with $\texttt{K}=1$ showing the most significant drops of 1.2 points across the metrics. Conversely, increasing the number of questions beyond $\texttt{K}=5$ yields small improvements, with $\texttt{K}=10$ showing the best relative gains of 1.6 points in F1.
The relatively small performance differences (within $\pm$2 points) indicate that our framework maintains robust performance across different $\texttt{K}$ values, with $\texttt{K}=5$ representing an effective balance between performance and computational efficiency.

\subsubsection{Using Small Language Models}\label{sec:4slm}

We investigate whether our KG-based contrastive reasoning methodology can enhance the performance of Small Language Models~(SLMs), as raised in question \textbf{RQ4}. We evaluate four SLMs: two with less than 600M parameters (SmolLM2~135M~\cite{allal2025smollm2} and Qwen3~0.6B~\cite{yang2025qwen3}) and two with less than 2B parameters (SmolLM2~1.7B~\cite{allal2025smollm2} and Qwen3~1.7B~\cite{yang2025qwen3}). We compare their F1 performance against Na\"{i}ve Claude 3.7 Sonnet and \namemodel{C3.7}.

\begin{figure}[htbp]
    \centering
    \includegraphics[width=0.925\linewidth]{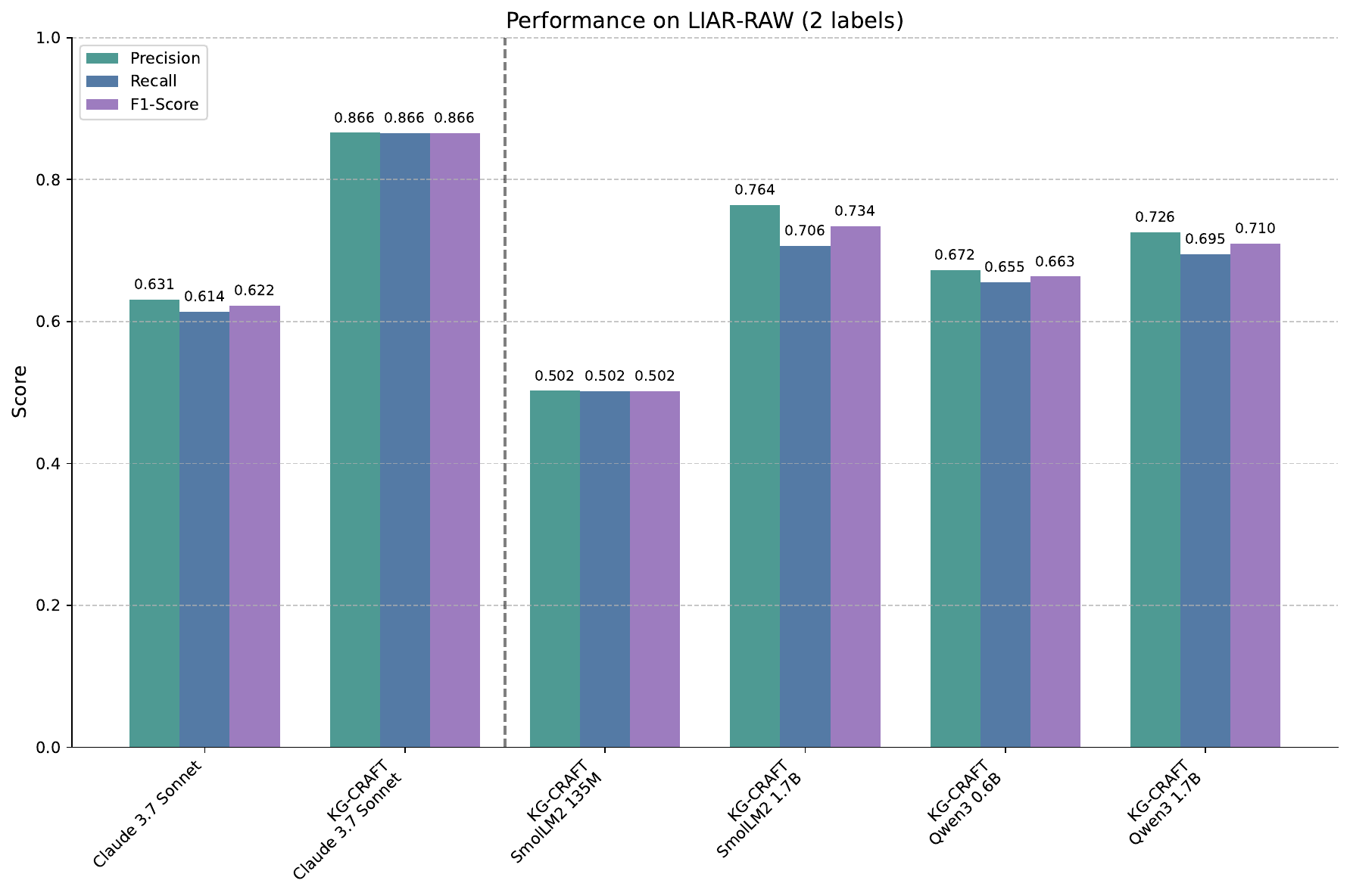}
    \caption{Performance comparison of Small Language Models incorporated within \name{} against larger models.}
    \label{fig:slm}
\end{figure}

The results in~\cref{fig:slm} show that our framework significantly enhances SLMs' fact-checking performance. Notably, SmolLM2~1.7B achieves an F1-Score of 73.40\%, substantially outperforming Na\"{i}ve Claude~3.7~Sonnet (62.22\%). 
Even the smaller models show promising results, with Qwen3~0.6B reaching an F1-Score of 66.34\%, which surpasses Na\"{i}ve Claude~3.7~Sonnet despite its much smaller size.

These findings suggest that \name{} effectively compensates for the limitations of smaller models by supplying them with relevant verification cues. Applying \name{} significantly narrows the performance gap between SLMs and larger models, indicating that the explicit use of contrastive questions grounded in extracted KGs enhances smaller models' competitiveness.

%% file: resources/tab-verification.tex
\begin{table*}[htbp]
\centering
\small
\caption{Fact-checking results (\%) on the RAWFC and LIAR-RAW datasets.}
\begin{tabular}{lcccccc}
\toprule
\multirow[b]{2}{*}{\textbf{Method}} & \multicolumn{3}{c}{\textbf{LIAR-RAW}} & \multicolumn{3}{c}{\textbf{RAWFC}} \\
\cmidrule(lr){2-4} \cmidrule(lr){5-7}
& \textbf{Pr} & \textbf{Re} & \textbf{F1} & \textbf{Pr} & \textbf{Re} & \textbf{F1} \\
\midrule
\addlinespace[0.5em]
\multicolumn{7}{l}{\textbf{Traditional}} \\
dEFEND~\cite{shu2019defend} & 23.09 & 18.56 & 17.51 & 44.93 & 43.26 & 44.07 \\
SBERT-FC~\cite{kotonya2020explainable} & 24.09 & 22.07 & 22.19 & 51.06 & 45.92 & 45.51 \\
GenFE~\cite{atanasova2020generating} & 28.01 & 26.16 & 26.49 & 44.92 & 44.74 & 44.43 \\
GenFE-MT~\cite{atanasova2020generating} & 18.55 & 19.90 & 15.15 & 45.64 & 45.27 & 45.08 \\
CofCED~\cite{yang2022coarse} & 29.48 & 29.55 & 28.93 & 52.99 & 50.99 & 51.07 \\
EExpFND~\cite{Wang2025eexpfnd} & 29.91 & 29.93 & 29.58 & 54.43 & 53.49 & 53.61 \\
\midrule
\addlinespace[0.5em]
\multicolumn{7}{l}{\textbf{Na\"{\i}ve LLM}} \\
Llama 2 7B~\cite{wang2024explainable} & 17.11 & 17.37 & 15.14 & 37.30 & 38.03 & 36.77 \\
ChatGPT 3.5 Turbo~\cite{wang2024explainable} & 25.41 & 27.33 & 25.11 & 47.72 & 48.62 & 44.43 \\
Claude 3.5 Sonnet & 29.25 & 27.83 & 28.52 & 54.85 & 55.04 & 54.94 \\
Claude 3.7 Sonnet & 28.26 & 26.63 & 27.42 &	55.90 &	57.12 &	56.50 \\
Llama 3.3 70B & 47.21 & 23.52 & 31.40 &	53.16 &	55.04 &	54.08 \\
\hdashline[0.5pt/2pt]
\addlinespace[0.5em]
\multicolumn{7}{l}{\textbf{Specialised LLM}} \\
FactLLAMA~\cite{cheung2023factllama} & 32.32 & 31.57 & 29.98 & 53.76 & 54.00 & 53.76 \\
FactLLAMA\textsubscript{know}~\cite{cheung2023factllama} & 32.46 & 32.05 & 30.44 & 56.11 & 55.50 & 55.65 \\
L-Defense\textsubscript{LLAMA2}~\cite{wang2024explainable} & 31.63 & 31.71 & 31.40 & 60.95 & 61.00 & 60.12 \\
L-Defense\textsubscript{ChatGPT}~\cite{wang2024explainable} & 30.55 & 32.20 & 30.53 & 61.72 & 61.01 & 61.20 \\
DelphiAgent\textsubscript{gpt-4o-mini}~\cite{xiong2025delphi} & 32.79 & 22.33 & 26.03 & 68.53 & 63.95 & 64.68 \\
DelphiAgent\textsubscript{gpt-4o}~\cite{xiong2025delphi} & 31.33 & 28.36 & 28.36 & 68.05 & 68.03 & 68.04 \\
\midrule
\addlinespace[0.5em]
\multicolumn{7}{l}{\textbf{\name{} (ours)}} \\
\namemodel{C3.5} & 62.70 &	59.38 &	60.99 &	67.10 &	66.25 &	66.67 \\
\namemodel{C3.7} & \textbf{73.92} &	\textbf{\underline{70.86}} & \textbf{72.36} & \textbf{69.37} & \textbf{68.59} &	\textbf{68.98} \\
\namemodel{L3.3} & \textbf{\underline{77.38}} & \textbf{70.67} & \textbf{\underline{73.87}} & \textbf{\underline{81.63}} & \textbf{\underline{81.53}} & \textbf{\underline{81.58}} \\
\bottomrule
\end{tabular}
\begin{tablenotes}
\centering
\item \textbf{Note:} The \textbf{\underline{best}} and \textbf{second best} results are highlighted across each dataset and metric. \name{} results are significantly better than their Na\"{i}ve LLM counterparts baseline with $p < 0.01$.
\end{tablenotes}
\label{tab:verification}
\end{table*}

%% file: secs/5-conclusion.tex
This paper presents \name{}, a novel method for claim veracity classification via knowledge graph-based contrastive reasoning with LLMs. 
Our results demonstrate that \name{} outperforms state-of-the-art methods on two real-world datasets (LIAR-RAW and RAWFC). 
Our empirical analysis indicates that the contrastive reasoning method generates contextually relevant and evidence-based information that aids in assessing claim veracity. 
We investigate the effect of the number of contrastive questions used in the process, finding that more questions lead to better results, but even small numbers lead to competitive performance. 
Even with Small Language Models, our approach proves competitive with LLM baselines. 
For future work, we propose an extended evaluation with other domains and the generation of explanations for full AFC based on contrastive answer summaries.

%% file: secs/6-limitations.tex
Whilst our results demonstrate the effectiveness of the proposed framework, some limitations remain. First, we do not qualitatively verify the intermediate components of our pipeline, such as the knowledge graph construction step, which is particularly sensitive and central to the method's overall performance. Additionally, we rely on a fixed set of LLMs for the intermediate tasks, though alternative models or fine-tuned approaches could potentially yield improved results. Our evaluation is limited to two datasets; however, these are widely adopted benchmarks in the fact-checking literature and provide a meaningful basis for comparison. We also acknowledge the reliance on expensive LLMs throughout the process, which may not be accessible to all users. However, our results also demonstrate the potential of the proposed framework to enhance the performance of smaller language models that require significantly fewer computational resources. 

Although we claim to improve the results of Specialised LLMs-based AFC, we are aware of a non-negligible potential threat to the validity of our results: the use of different families of LLMs in previous and in our work. The results may therefore be influenced by variations in model capabilities, and different outcomes could emerge if their LLMs or ours were replaced. However, we were unable to modify the original systems' LLMs or employ legacy models in our experiments. This highlights a broader challenge for the community: ensuring reproducibility and fair comparison in an ecosystem where new and more capable LLMs are continuously emerging. Nonetheless, we presented ablation studies of \name{} with LLMs of limited size (SLMs), showing that it significantly enhances their performance.

Furthermore, all experiments are conducted in the English language. Although our method is designed to be language-agnostic, performance may vary across languages due to potential limitations in intermediate components such as entity linking, relation extraction, or question generation. Future work should explore broader language coverage, dataset diversity, and deeper analysis of intermediate outputs.

%% file: secs/a5-background.tex
\section{Background}\label{sec:abg}

\subsection{Key Concepts}

\paragraph{Contrastive Explanations and Reasoning\label{sec:a2contrastivekc}}
The principle of contrastive explanations is rooted on answering counterfactual why-question by comparing the actual outcome with hypothetical alternatives~\cite{lipton1990contrastive,guidotti2024counterfactual,verma2024counterfactual}. The contrastiveness presupposes that an explanation answers ``Why did $P$ happen?'' in terms of ``Why did $P$ happen rather than $Q$?'', where $P$ is an observed event and $Q$ represents alternative hypotheses~\cite{stepin2021survey}. This approach ensures that explanations provide comprehensive information by distinguishing the chosen outcome from a set of contrastive hypothetical alternatives, establishing a minimum criterion where explanations must demonstrate why the observed event was more probable than its alternatives.

In short, contrastive explanations aim to answer why-questions by comparing an observed outcome $P$ with counterfactual alternatives $Q$. Rather than just explaining why $P$ occurred, they focus on why $P$ happened instead of $Q$, highlighting why the chosen outcome is more plausible.



\paragraph{Knowledge Graph Construction Using LLMs}
A knowledge graph~(KG) represents information as a graph structure, where nodes are \emph{entities} connected by \emph{relations}. Formally, $\mathcal{G} = (\mathcal{E}, \mathcal{R}, \mathcal{T}, \mathbb{C})$, where $\mathcal{T}$ contains triples $(h, r, t)$ with entities $\{h, t\} \in \mathcal{E}$ and relations $r \in \mathcal{R}$.

The use of LLMs to automate KG construction has been a theme of several works~\cite{chen2023autokg,zhang2024extract,pan2024kgllm}, specifically due to their performance on extracting entities and relations~\cite{zhu2024llms}.

\subsection{Related Work}

\textbf{Contrastive Explanations and Reasoning\label{sec:a2contrastiverw}}
\citet{jacovi2021contrastive} applied contrastive explanations by projecting inputs into a latent space that captures only features distinguishing potential decisions, enabling models to better identify which aspects support or contradict specific predictions. 
\citet{paranjape2021contrastive} extended this concept to commonsense reasoning tasks, showing that pre-trained language models can generate contrastive explanations that highlight key differences between alternatives, improving both performance and explanation faithfulness.

These results show the effectiveness of a contrastive approach to reasoning and indicate its potential for claim verification. 

\paragraph{Non-generative Automated Fact-checking\label{sec:a2ml}}
AFC typically relies on text embeddings to represent claims and supporting documents. 
A notable example is dEFEND, which introduced a sentence-comment co-attention network to jointly analyse news content and user comments~\cite{shu2019defend}. 
\citet{kotonya2020explainable} developed SBERT-FC, introducing the PUBHEALTH dataset to expand the thematic focus on public health and provide extended explainability analysis. 
GenFE and GenFE-MT \cite{atanasova2020generating} added joint optimisation of veracity prediction and explanation generation, showing improved classification and output quality. 

CofCED~\cite{yang2022coarse} introduced a novel hierarchical architecture, combining an encoder with cascaded evidence selectors to process reports from multiple sources for claim verification and explanation generation. 
\citet{tillman2022icwsm} integrated knowledge bases into pre-trained models and showed that using Wikidata improves accuracy, particularly for political claims where the knowledge base is timely and relevant. More recently, \citet{qudus2025kgfc} surveyed the use of KGs for AFC.


\paragraph{Fact-checking Using LLMs\label{sec:a2llm}}
The emergence of LLMs and their strong performance across domains has led to a new set of approaches that place them at the core of AFC. 
Whilst LLMs can give answers to factual questions, their reliability is limited by the extent of their training data and their tendency to hallucinate factual statements~\cite{wang2024factuality,augenstein2024factuality}. Several frameworks have been proposed to address the challenges of reliability and transparency in LLM-based fact-checking. 

Early work in this direction includes FactLLaMA~\cite{cheung2023factllama}, which combined instruction-following with external evidence retrieval and demonstrated that augmenting LLMs with external knowledge sources improves fact-checking accuracy, particularly for claims related to recent events. 
IKA~\cite{guo2023interpretable} builds a graph of positive and negative examples from labelled data to enhance fact-checking and explanation capabilities. 
\citet{liu2024teller} introduced TELLER, a dual-system framework that emphasises trustworthiness through the integration of human expertise and LLM capabilities. 
Similarly, \citet{wang2024explainable} developed a defence-based framework that addresses the limitations of uncensored crowd wisdom by splitting evidence into competing narratives and leveraging LLMs for reasoned verification. More recently, CorXFact~\cite{tan2025improving} proposes simulating human fact-checking principles by analysing claim-evidence correlations.

A popular approach to improving LLM fact-checking is Retrieval-Augmented Generation~(RAG). 
Examples include the use of basic RAG-based architectures~\cite{singal2024evidence}, RAG enhanced with fine-grained feedback mechanisms for optimising the retrieval task~\cite{zhang2024reinforcement}, and more advanced RAG-based architectures focused on evidence retrieval and contrastive argument synthesis~\cite{yue2024retrieval}.

Other recent approaches focus on specific aspects of fact-checking. 
For instance, FIRE~\cite{xie2025fire} proposes an iterative approach to fact-checking claims aiming to improve scalability and efficiency and \citet{meng2025detecting} address the challenge of zero-day manipulated content through real-time contextual information retrieval.

%% file: secs/a1-implementation.tex
\section{Implementation Details}\label{sec:aid}
This section presents information about our experimental setup. We first detail the language models employed and their configuration settings, followed by the description and statistics of the datasets used in our evaluation (\cref{sec:4es}). Additionally, we present our modification to the \textit{AlignScore} metric~\cite{zha2023alignscore} (\cref{sec:4llmcq}).

\begin{table*}[!t]
\centering
\caption{Overview of language models used in our experiments, including model providers and custom hyperparameter settings.}
\label{tab:models}
\resizebox{\textwidth}{!}{
\begin{tabular}{llll}
\toprule
\textbf{Model Name} & \textbf{Model Provider} & \textbf{Model Identifier} & \textbf{Custom Hyperparameters} \\
\midrule
Claude 3 Haiku & Amazon Bedrock & anthropic.claude-3-haiku-20240307-v1:0 & \texttt{temperature = 0.0} \\
Claude 3.5 Haiku & Amazon Bedrock & us.anthropic.claude-3-5-haiku-20241022-v1:0 & \texttt{temperature = 0.0} \\
Claude 3.5 Sonnet & Amazon Bedrock & us.anthropic.claude-3-5-sonnet-20241022-v2:0 & \texttt{temperature = 0.0} \\
Claude 3.7 Sonnet & Amazon Bedrock & us.anthropic.claude-3-7-sonnet-20250219-v1:0 & \texttt{temperature = 0.0} \\
Llama 3.3 70B & Amazon Bedrock & us.meta.llama3-3-70b-instruct-v1:0 & \texttt{temperature = 0.0} \\
\addlinespace[0.25em]
\multirow{2}{*}{SmolLM2 135M} & \multirow{2}{*}{Hugging Face} & \multirow{2}{*}{HuggingFaceTB/SmolLM2-135M-Instruct} & \begin{tabular}[c]{@{}l@{}}\texttt{max\_length = 8192}\\\texttt{max\_new\_tokens = 128}\end{tabular} \\
\addlinespace[0.25em]
\multirow{2}{*}{SmolLM2 1.7B} & \multirow{2}{*}{Hugging Face} & \multirow{2}{*}{HuggingFaceTB/SmolLM2-1.7B-Instruct} & \begin{tabular}[c]{@{}l@{}}\texttt{max\_length = 8192}\\\texttt{max\_new\_tokens = 128}\end{tabular} \\
\addlinespace[0.25em]
\multirow{2}{*}{Qwen3 0.6B} & \multirow{2}{*}{Hugging Face} & \multirow{2}{*}{Qwen/Qwen3-0.6B} & \begin{tabular}[c]{@{}l@{}}\texttt{max\_length = 8192}\\\texttt{max\_new\_tokens = 32768}\end{tabular} \\
\addlinespace[0.25em]
\multirow{2}{*}{Qwen3 1.7B} & \multirow{2}{*}{Hugging Face} & \multirow{2}{*}{Qwen/Qwen3-1.7B} & \begin{tabular}[c]{@{}l@{}}\texttt{max\_length = 8192}\\\texttt{max\_new\_tokens = 32768}\end{tabular} \\
\bottomrule
\end{tabular}
}
\end{table*}

\subsection{Models Settings}

We evaluate our framework using a diverse set of language models, ranging from large-scale models available through Amazon Bedrock\footnote{\url{https://aws.amazon.com/bedrock/}} to smaller, open-source alternatives from Hugging Face\footnote{\url{https://huggingface.co/}} (Hugging Face models were deployed using Amazon SageMaker AI\footnote{\url{https://aws.amazon.com/sagemaker-ai/}}). Table~\ref{tab:models} presents the complete list of models used in our experiments, along with their providers and custom hyperparameters. For all Amazon Bedrock models, we set the temperature to 0.0 to ensure deterministic outputs. For the Hugging Face models (SmolLM2 and Qwen3 families), we configure specific maximum length and token generation parameters to accommodate our fact-checking pipeline requirements.

\subsection{Datasets}\label{sec:ads}

We evaluated the proposed approach on two fact-checking real-world datasets: LIAR-RAW and RAWFC (statistics shown in Table~\ref{tab:dsstats}). LIAR-RAW~\cite{yang2022coarse} extends LIAR-PLUS~\cite{alhindi2018evidence} and contains 12,590 claims from Politifact\footnote{\url{https://www.politifact.com/}} associated to six veracity classes. RAWFC~\cite{yang2022coarse} comprises of 2,012 claims collected from Snopes\footnote{\url{https://www.snopes.com/}} with three veracity classes. Both datasets have their claims split into three sets: train, test, and validation, containing, respectively, 80\%, 10\%, and 10\% of the claims. LIAR-RAW and RAWFC are used in \cref{sec:4cve,sec:4llmcq}. A modified version of LIAR-RAW, called \textit{LAIR-RAW (2 labels)}, used in~\cref{sec:4kcq,sec:4slm}, is also described in the table. This modification maps the original six labels into two classes: \{\texttt{pants-fire}, \texttt{false}, \texttt{barely-true}\} as \texttt{false}, and \{\texttt{half-true}, \texttt{mostly-true}, \texttt{true}\} as \texttt{true}, whilst preserving the original claims split.

\begin{table}[H]
\centering
\caption{Statistics of the LIAR-RAW and RAWFC datasets. $\mid\mathcal{C}\mid_{ALL}$ denotes the total number of claims, and $\mid\mathcal{R}\mid_{avg}$ represents the average number of reports per claim.}
\begin{tabular}{l|l|r}
    \hline
    \textbf{Dataset} & \textbf{Statistics} & \textbf{Value} \\
    \hline
    \multirow{8}{*}{\rotatebox{90}{\centering LIAR-RAW}} & \texttt{pants-fire} & 1,013 \\
    & \texttt{false} & 2,466 \\
    & \texttt{barely-true} & 2,057 \\
    & \texttt{half-true} & 2,594 \\
    & \texttt{mostly-true} & 2,439 \\
    & \texttt{true} & 2,021 \\
    & $\mid\mathcal{C}\mid_{ALL}$ & 12,590 \\
    & $\mid\mathcal{R}\mid_{avg}$ & 12.3 \\
    \hline
    \multirow{5}{*}{\rotatebox{90}{\centering RAWFC}}& \texttt{false} & 646 \\
    & \texttt{half} & 671 \\
    & \texttt{true} & 695 \\
    & $\mid\mathcal{C}\mid_{ALL}$ & 2,012 \\
    & $\mid\mathcal{R}\mid_{avg}$ & 21.0 \\
    \hline
    \rotatebox{0}{\centering{LIAR-RAW}} & \texttt{false} & 5,536 \\
    \rotatebox{0}{\centering{(2 labels)}} & \texttt{true} & 7,054 \\
    \hline
\end{tabular}
\label{tab:dsstats}
\end{table}

\subsection{Weighted AlignScore and RQUGE\label{sec:awalignscore}}

As described in~\cref{sec:4llmcq}, \emph{AlignScore} is a metric based on a general function of information alignment to perform automatic factual consistency evaluation of text pairs~\cite{zha2023alignscore}. We extended AlignScore to weight its results by the distance of the predicted claim veracity class from the ground truth claim class.

For that, first, based on the semantic distance between classes, we progressively assign each class an integer numerical value ranging from 1 ($y_{min}$) to $\mid C \mid$ ($y_{max}$), where $C$ is the set of possible labels. For instance, in the case of LIAR-RAW, we assigned each class an integer numerical value ranging from 1 (\texttt{pants-fire}) to 6 (\texttt{true}). After, we measure the mean squared error~(MSE) between the predicted class. Finally, as depicted in \cref{eq:walignscore}), we weight the macro AlignScore for the pair answer summarisation text and original claim using the obtained MSE.
\begin{equation}\label{eq:walignscore}
\small    \text{AlignScore}_{w} = \left(1 - \frac{(\mathcal{V}_\mathcal{C} - y)^2}{(y_{\text{max}} - y_{\text{min}})^2}\right) \times AlignScore(A_{\mathcal{C}}, \mathcal{C})
\end{equation}
With this, we penalise the alignment scores of answer summary $A_\mathcal{C}$ and claim $\mathcal{C}$ where the fact-checking model prediction $\mathcal{V}_\mathcal{C}$ does not match the expected class $y$ based on the answer summary, while considering that predicting closer classes is better than predicting distinct classes.

\subsection{Weighted RQUGE\label{sec:awrquge}}
As described in~\cref{sec:4llmcq}, \emph{RQUGE} scores the acceptability of a generated question given a context and an answer~\cite{mohammadshahi2023rquge}. Let $q_i \in Q^\texttt{K}_{ranked}$ be a (formulated contrastive) question for claim $\mathcal{C}$ (context) with corresponding answer $a_i \in \tilde{A}$; denote the macro score over a set of $K$ questions as

\begin{equation*}
\begin{aligned}
\text{RQUGE}_{\text{macro}}\big(\mathcal{C}\big)\\
=~& \frac{1}{K}\sum_{k=1}^{K}\text{RQUGE}(q_k,\mathcal{C},a_k) .
\end{aligned}
\end{equation*}

Analogously to \cref{sec:awalignscore}, we weight RQUGE by the distance between the model’s predicted veracity class $\mathcal{V}_{\mathcal{C}}$ and the gold label $y$. Using the same class mapping $y_{\min}\!\to\!y_{\max}$ and mean-squared error term, the per-claim weight is
\[
w_{\mathcal{C}} = 1-\frac{(\mathcal{V}_{\mathcal{C}}-y)^2}{(y_{\text{max}}-y_{\text{min}})^2}.
\]
We define the weighted RQUGE for a single question and its macro form as:
\begin{equation}\label{eq:wrquge}
\small
\text{RQUGE}_{w}(\mathcal{C}) = w_{\mathcal{C}}\times \text{RQUGE}_{\text{macro}}(\mathcal{C}).
\end{equation}
This penalises question-quality scores when the verifier’s prediction $\mathcal{V}_{\mathcal{C}}$ deviates from $y$, while granting higher credit to questions associated with \emph{closer} (semantically adjacent) class predictions.

%% file: secs/a3-exp.tex
\section{Additional Experiments}\label{sec:aexp}

\begin{table*}[!h]
\centering
\small
\caption{Comparative performance analysis (\%) of KG-CRAFT and its key components on LIAR-RAW and RAWFC datasets.}
\begin{tabular}{lcccccc}
\toprule
\multirow[b]{2}{*}{\textbf{Method}} & \multicolumn{3}{c}{\textbf{LIAR-RAW}} & \multicolumn{3}{c}{\textbf{RAWFC}} \\
\cmidrule(lr){2-4} \cmidrule(lr){5-7}
& \textbf{Pr} & \textbf{Re} & \textbf{F1} & \textbf{Pr} & \textbf{Re} & \textbf{F1} \\
\midrule
\multicolumn{7}{l}{\textbf{\name{} (ours)}} \\
\namemodel{C3.5} & 62.70 &	59.38 &	60.99 &	67.10 &	66.25 &	66.67 \\
\namemodel{C3.7} & \textbf{73.92} &	\textbf{\underline{70.86}} & \textbf{72.36} & \textbf{69.37} & \textbf{68.59} &	\textbf{68.98} \\
\namemodel{L3.3} & \textbf{\underline{77.38}} & \textbf{70.67} & \textbf{\underline{73.87}} & \textbf{\underline{81.63}} & \textbf{\underline{81.53}} & \textbf{\underline{81.58}} \\

\midrule
\addlinespace[0.5em]

\multicolumn{7}{l}{\textbf{Na\"{\i}ve LLM}} \\
Claude 3.5 Sonnet & 29.25 & 27.83 & 28.52 & 54.85 & 55.04 & 54.94 \\
Claude 3.7 Sonnet & 28.26 & 26.63 & 27.42 &	55.90 &	57.12 &	56.50 \\
Llama 3.3 70B & 47.21 & 23.52 & 31.40 &	53.16 &	55.04 &	54.08 \\
\hdashline[0.5pt/2pt]
\addlinespace[0.5em]
\multicolumn{7}{l}{\textbf{LLM with KG augmentation (no contrastive reasoning)}} \\
Claude 3.5 Sonnet & 39.96 & 37.52 & 38.70 & 56.73 & 56.51 & 56.62 \\
Claude 3.7 Sonnet & 60.70 & 56.42 & 58.48 &	68.27 &	66.09 &	67.16 \\
Llama 3.3 70B & 61.50 & 38.11 & 47.06 &	60.64 &	58.54 &	59.57 \\
\hdashline[0.5pt/2pt]
\addlinespace[0.5em]
\multicolumn{7}{l}{\textbf{LLM-based contrastive questions (no KG)}} \\
Claude 3.5 Sonnet & 34.54 & 23.25 & 27.79 & 65.40 & 66.56 & 65.97 \\
Claude 3.7 Sonnet & 30.20 & 27.05 & 28.54 & 67.65 & 68.57 & 68.11 \\
Llama 3.3 70B & 42.18 & 22.90 & 29.68 & 64.98 & 64.99 & 64.98 \\
\bottomrule
\end{tabular}
\begin{tablenotes}
\centering
\item \textbf{Note:} The \textbf{\underline{best}} and \textbf{second best} results are highlighted across each dataset and metric. \textit{\name{}} and \textit{Na\"{\i}ve LLM} results are same in~\cref{tab:verification}.
\end{tablenotes}
\label{tab:extraverification}
\end{table*}

\subsection{Ablation Study\label{sec:aas}}
To evaluate the impact of \name{}'s components and validate its effectiveness as a complete framework, we conducted a comprehensive ablation study on the LIAR-RAW and RAWFC datasets. Our analysis compares the full \name{} framework (proposed in~\cref{sec:3kgcraft} and results discussed in~\cref{sec:4cve}) against three key architectural variations: Na\"ive LLM, an LLM augmented with only the knowledge graph~(KG), and an LLM that generates contrastive questions without using the KG structure (presented and discussed in~\cref{sec:4llmcq}). The results, presented in Table~\ref{tab:extraverification}, provide a detailed breakdown of each component's contribution to the overall performance.

\paragraph{Na\"ive LLM} The baselines Na\"ive LLM section (presented in~\cref{sec:4es} and results discussed in~\cref{sec:4cve}) shows the performance of aforementioned backbone LLMs when tasked with fact-checking using only the claim and its associated reports. This approach lacks any structured reasoning or pre-processing of the reports. The results for these models are notably lower than for \name{}, confirming the need for an enhanced reasoning mechanism.

\paragraph{LLM with KG augmentation (no contrastive reasoning)} This ablation evaluates the backbone models augmented of the knowledge graphs extracted from the claim and reports, but bypassing the contrastive reasoning phase. The knowledge graph~(KG) is provided as additional context to the LLM for veracity assessment. Comparing these results to the Na\"ive LLM baselines shows that augmenting the LLM with the extract (KG) does improve performance. However, the gains are marginal compared to the full \name{} framework, highlighting that the contrastive reasoning process is the primary driver of the significant performance increase. For instance, on the LIAR-RAW dataset, Llama 3.3 70B shows a 15pp increase in F1-score with KG augmentation, but the full KG-CRAFT framework provides a more substantial 42pp gain compared to the na\"ive baseline.

\paragraph{LLM-based Contrastive Questions (No KG)} This ablation directly addresses RQ2~(\cref{sec:4as}) by replacing the KG-based question formulation with questions generated purely by the LLM using a few-shot prompt (presented and discussed in~\cref{sec:4llmcq}). The results indicate that this approach is less effective than our KG-based method. The F1-scores are significantly lower, for instance, with Llama 3.3 70B achieving only a 29.68\% F1-score on LIAR-RAW, compared to the 73.87\% F1-score of the full \name{} framework. This suggests that LLMs, when prompted to generate their own contrastive questions, often fail to create questions that are both contextually relevant and aligned with the provided reports, resulting in overall lower information alignment~(AlignScore) between the answer summaries and the original claims and overall lower question quality~(RQUGE) between formulated questions, generated answers, and the original claims~(\cref{sec:4llmcq}). This finding reinforces the value of our knowledge graph approach for producing evidence-based questions.

\paragraph{KG-CRAFT (ours)} The results of \name{} framework (proposed in~\cref{sec:3kgcraft} and results discussed in~\cref{sec:4cve}), which combines all proposed components, consistently outperform all other variations, achieving a new state-of-the-art on both datasets~(\cref{sec:4cve}). This confirms that the combination of knowledge graph extraction with the proposed contrastive reasoning significantly enhances LLMs' fact-checking abilities.

\subsection{Evaluation with Other Datasets\label{sec:aeds}}

We further examine whether \name{} generalises to scenarios where claims are accompanied by fewer and shorter reports by evaluating on two benchmarks: \textbf{SciFact}~(scientific abstracts) and \textbf{PubHealth}~(health and policy claims).

\subsubsection{Experimental Settings}

\paragraph{Datasets}
SciFact dataset~\cite{wadden2020scifact} targets scientific claim verification from research paper abstracts.  We used its validation set (also referred to as \textit{dev} set), retaining claims with complete supporting or refuting evidence (label classes: \texttt{supports} and \texttt{refutes}). PubHealth~\cite{kotonya2020explainable} covers health-related and public policy claims. We use the test split, selecting claims that include the dataset’s supporting context (label classes: \texttt{true}, \texttt{false}, \texttt{mixture}). In both settings, we strictly use the reports provided by each dataset, \ie{}, no external retrieval (also known as gold evidence), so observed differences reflect reasoning rather than retrieval. Full results are in Tables~\cref{tab:scifact}~and~\cref{tab:pubhealth_scifact}.

\paragraph{Comparisons}
The performance of \name{} is benchmarked against seven competing methods, which include encoder-based classifiers, graph models, program-style pipelines, and LLM-based approaches.
MLA~(RoBERTa)~\cite{kruengkrai2021mla} proposes a sequence inference model which uses self-attention at both the token and sentence levels to capture information with a pre-LM encoder (RoBERTa). It also feeds static positional encodings into its multi-head attention mechanism.
MULTIVERS~\cite{wadden2022multivers} leverages a multi-task learning approach aiming at multi-evidence scientific verification and rationale extraction from abstracts.
ProgramFC~\cite{pan2023programfc} formulates fact verification as a programmatic pipeline with modular steps for evidence usage and decision making.
PACAR~\cite{zhao2024pacar} is a prompting-based approach that aggregates evidence with consistency-oriented reasoning.
GraphFC~\cite{huang2025graphfc} encodes the report structure with an explicit graph and graph reasoning components.
CO-GAT (ELECTRA)~\cite{lan2025cogat} applies graph attention over scientific evidence with a pre-LLM encoder (ELECTRA).
GraphCheck~\cite{chen2025graphcheck} is a recent LLM-based verifier that incorporates lightweight graph signals and instruction-style prompting.
We report \name{} with two backbones: \name{}\textsubscript{C3.5} and \name{}\textsubscript{L3.3}, with the same settings as reported in our main experiments~(\cref{sec:4es}).

\paragraph{Evaluation Metrics} As in our main experiments (\cref{sec:4es}), we report precision~(Pr), recall~(Re), and F1-score~(F1) results. In addition, to compare with GraphCheck, we report balanced accuracy~(BAcc). For all metrics, higher values indicate better performance.

\begin{table*}[!t]
\centering
\small
\caption{Performance comparison on the SciFact dataset (\%).}
\begin{tabular}{lccc}
\toprule
\textbf{Method} & \textbf{Pr} & \textbf{Re} & \textbf{F1} \\
\midrule
\addlinespace[0.5em]
\multicolumn{4}{l}{\textbf{Previous Methods}} \\
MLA (RoBERTa)~\cite{kruengkrai2021mla}$^a$ & \textbf{80.62} & 49.76 & 61.54 \\
MULTIVERS~\cite{wadden2022multivers} & 73.80 & 71.20 & 72.50 \\
ProgramFC~\cite{pan2023programfc}$^b$ & - & - & 71.82 \\
PACAR~\cite{zhao2024pacar} & - & - & 75.06 \\
GraphFC~\cite{huang2025graphfc} & - & - & \textbf{\underline{87.37}} \\
CO-GAT (ELECTRA)~\cite{lan2025cogat}$^d$ & 79.58 & 54.07 & 64.39 \\
\midrule
\addlinespace[0.5em]
\multicolumn{4}{l}{\textbf{KG-CRAFT (ours)}} \\
KG-CRAFT\textsubscript{C3.5} & 76.50 & \textbf{75.18} & 75.83 \\
KG-CRAFT\textsubscript{L3.3} & \textbf{\underline{84.58}} & \textbf{\underline{81.53}} & \textbf{83.03} \\
\bottomrule
\end{tabular}
\begin{tablenotes}
\centering
\item \textbf{Note:} The \textbf{\underline{best}} and \textbf{second best} results are highlighted across each dataset and metric. All reported results use the evidence provided by the dataset; thus, no external source is referenced. $^a$ Results taken from~\cite{lan2025cogat}. $^b$ Results taken from~\cite{zhao2024pacar}. $^c$ Results taken from~\cite{huang2025graphfc}. $^d$ CO-GAT (ELECTRA) reported results use the large model and abstract-level evidence settings. MULTIVERS reported results use the full and abstract-level evidence settings.
\end{tablenotes}
\label{tab:scifact}
\end{table*}


\begin{table*}[!t]
\centering
\small
\caption{Performance comparison on the PubHealth and SciFact datasets (\%).}
\begin{tabular}{lcccccccc}
\toprule
\multirow[b]{2}{*}{\textbf{Method}} & \multicolumn{4}{c}{\textbf{PubHealth}} & \multicolumn{4}{c}{\textbf{SciFact}} \\
\cmidrule(lr){2-5} \cmidrule(lr){6-9}
& \textbf{BAcc} & \textbf{Pr} & \textbf{Re} & \textbf{F1} & \textbf{BAcc} & \textbf{Pr} & \textbf{Re} & \textbf{F1} \\
\midrule
\addlinespace[0.5em]
GraphCheck\textsubscript{L3.3}~\cite{chen2025graphcheck} & 73.60 & - & - & - & \textbf{89.40} & - & - & - \\
GraphCheck\textsubscript{Qwen~72B}~\cite{chen2025graphcheck} & 71.70 & - & - & - & 86.40 & - & - & - \\
\midrule
\addlinespace[0.5em]
\multicolumn{9}{l}{\textbf{KG-CRAFT (ours)}} \\
KG-CRAFT\textsubscript{C3.5} & 61.02 & 76.50 & 75.18 & 75.83 & 75.17 & 76.50 & 75.18 & 75.83 \\
KG-CRAFT\textsubscript{L3.3} & \textbf{78.66} & 72.82 & 73.89 & 73.35 & 81.52 & 84.58 & 81.53 & 83.03 \\
\bottomrule
\end{tabular}
\begin{tablenotes}
\centering
\item \textbf{Note:} The \textbf{\underline{best}} results for the Balanced Accuracy~(BAcc) are highlighted across each dataset. KG-CRAFT Precision, Recall, and F1-score are reported for completeness; they are not being compared with GraphCheck results.
\end{tablenotes}
\label{tab:pubhealth_scifact}
\end{table*}

\subsubsection{Results}
On SciFact, \name{}\textsubscript{L3.3} obtains 83.03 F1 (Pr/Re: 84.58/81.53), outperforming five out of the six competing techniques (\eg{}, MULTIVERS 72.50 and ProgramFC 71.82; more than 10pp difference) whilst presenting competing results with the best performing method (GraphFC 87.37; less than 5pp)  (Table~\ref{tab:scifact}).
Also, in SciFact, GraphCheck\textsubscript{L3.3} obtained 89.40 BAcc, outperforming \name{}\textsubscript{L3.3} in 7.88pp (Table~\ref{tab:pubhealth_scifact}).
On PubHealth, \name{}\textsubscript{L3.3} achieves the best BAcc at 78.66, surpassing GraphCheck\textsubscript{L3.3} (73.60; 5.06 pp) and GraphCheck\textsubscript{Qwen~72B} (71.70; 6.96 pp) (Table~\ref{tab:pubhealth_scifact}).
Whilst \name{}\textsubscript{C3.5} is consistently weaker than \name{}\textsubscript{L3.3} results, where reported for completeness.

In both datasets, reports associated with each claim are shorter, we observe less diverse extracted KGs, \ie{}, fewer entities and relations, which directly constrains the space of type-consistent substitutions and thus the capacity to formulate diverse contrastive questions. This likely contributes to the residual gap to GraphFC on SciFact, despite strong overall results and cross-domain robustness evidenced by PubHealth.

%% file: secs/a2-prompts.tex
\section{Prompt Engineering}\label{sec:ape}

This section presents the five prompts used in the scope of this work: Knowledge Graph Extraction (\cref{sec:3kge}), Contrastive Question Answer Generation (\cref{sec:3cqag}), Answer Summarisation (\cref{sec:3as}), Claim Veracity Verification (\cref{sec:3cv}), and Contrastive Question Formulation (\cref{sec:4llmcq}). For each prompt, we provide the template.

\subsection{Knowledge Graph Extraction Prompt}\label{sec:akgep}
The following prompt describes our phased approach to knowledge graph extraction, where the LLM is guided to sequentially identify entities ($\mathcal{E}$), assign their classes ($\mathbb{C}$), and establish relationships ($\mathcal{R}$) between them to construct the input knowledge graph $\mathcal{G}$.

\promptbox[footnotesize]{Knowledge Graph Extraction Prompt}{
You are a top-tier algorithm designed for extracting information in structured formats to build a knowledge graph.
\\
Knowledge graphs consist of a set of triples. Each triple contains two entities (subject and object) and one relation that connects these subject and object.
\\
Try to capture as much information from the text as possible without sacrificing accuracy. Do not add any information that is not explicitly mentioned in the text.
\\
This is the process to extract information and build a knowledge graph:
\\
1. Extract nodes [...]
\\
2. Label nodes [...]
\\
3. Extract relationships [...]
\\
Compliance criteria: [...]
\\
Text: \textit{\{claim or report\}}\\
}

\subsection{Contrastive Question Answer Generation Prompt\label{sec:aagp}}
The following prompt instructs the LLM to generate answers to contrastive questions ($\mathcal{Q}_{ranked}^\texttt{K}$) by analysing claim-associated reports ($\mathcal{R}_\mathcal{C}$), ensuring responses are grounded in evidence whilst maintaining traceability between claims, reports, and questions.

\promptbox[footnotesize]{Contrastive Question\\Answer Generation Prompt}{
You are an expert answering questions based only on the provided context.\\~\\
\#\# Task:\\
Using the context provided and being aware of the claim, answer the question regarding the claim aiming to fact-check it. Limit your answer to 200 words at most.\\~\\
\#\# Desired Outcome:\\
- Base the concise answer strictly on the context.\\
- Present the information neutrally, without judging or labeling the claim.\\
- Do not re-state the claim in the answer.\\
- Write in continuous prose (no lists, bullet points, or meta-commentary).\\
- Limit your answer to 200 words at most.\\~\\
\#\# Input:\\
* Context: \textit{\{context\}}\\
* Claim: \textit{\{claim\}}\\
* Question: \textit{\{contrastive question\}}\\~\\
\#\# Output:
}

\subsection{Answer Summarisation Prompt\label{sec:aasp}}
The following prompt instructs the LLM to generate a concise summary ($A_{\mathcal{C}}$) from the claim ($\mathcal{C}$) and its associated question-answer pairs ($\mathcal{Q}_{ranked}^\texttt{K}$, $\tilde{\mathcal{A}}$), emphasizing key contrasting elements while abstracting non-essential information.

\promptbox[footnotesize]{Answer Summarisation Prompt}
{
You are an expert writing summarizing information from pairs of question and answer.\\~\\
\#\# Task:\\
Your task is to generate an one paragraph summary of the information based on given pairs of question and answer.\\~\\
\#\# Desired Outcome:\\
- A one paragraph summary of the information contained in the question and answer.\\
- Present the information neutrally, without judging or labeling the claim.\\
- Ensure that the summary is clear and accurately based on the provided context.\\
- Write in continuous prose (no lists, bullet points, or meta-commentary).\\
- Do not add any utterances (for example "Here are" statements) to the final answer.\\
- Limit your answer to 200 words at most.\\~\\
\#\# Input:\\
* Question 1: \textit{\{contrastive question\}}\\
* Answer 1: \textit{\{contrastive question answer\}}\\
* [...]\\~\\
\#\# Output:
}

\subsection{Claim Veracity Verification Prompt\label{sec:acvvp}}

The following prompt ($p_\text{cv}$) instructs the LLM to determine claim veracity ($\mathcal{V}_\mathcal{C}$) by analysing the original claim ($\mathcal{C}$) against the distilled evidence summary ($A_{\mathcal{C}}$), using predefined veracity labels and their descriptions.


\promptbox[footnotesize]{Claim Veracity Verification Prompt}{
You are an expert fact-checking claims based solely on the provided context of the claim.\\~\\
\#\# Task:\\
Your task is to categorize the claim based only on the context as:\\
-  \textit{\{veracity labels\}}\\~\\
\#\# Desired Outcome:\\
- The veracity of the claim based on the context provided.\\
- Your response must be only one of the he above options. Do not include any other text.\\~\\
\#\# Input:\\
* Context: \textit{\{context\}}\\
* Claim: \textit{\{claim\}}\\~\\
\#\# Output:
}

\subsection{Contrastive Question Formulation Prompt\label{sec:cqfp}}

The following prompt guides the LLM to generate contrastive questions directly from claims and reports, serving as an alternative to the knowledge graph-based approach for our ablation study.

\promptbox[footnotesize]{Contrastive Question Formulation Prompt}{
You are an expert writing analyzing a given claim and generating contrastive questions based on given context. Your task is to generate contrastive questions of given claim based on given context.\\~\\
\#\# Desired Outcome:\\
- Create contrastive questions of an input claim based on given context.\\
- Present a list of five contrastive questions.\\
- Do not add any utterances (for example "Here are" statements) to the final answer.\\~\\
\#\# Example Prompt:\\
* Claim: \textit{\{claim example\}}\\
* Context: \textit{\{reports examples\}}\\~\\
\#\# Example Output:\\
''''''\\
\textit{\{contrastive questions examples\}}\\
''''''\\~\\
\#\# Additional Notes:\\
- Ensure that the contrastive questions are clear and accurately contrasts the claim based on the provided context.\\
- Maintain a consistent and readable format for the output.\\
- Ensure that the output is only the contrastive questions, no other additional text or utterances.\\~\\
\#\# Input:\\
* Claim: \textit{\{claim\}}\\
* Context: \textit{\{reports\}}\\~\\
\#\# Output:
}